%% file: main.tex
\documentclass[letterpaper]{article} 
\usepackage{aaai25}  
\usepackage{times}  
\usepackage{helvet}  
\usepackage{courier}  
\usepackage[hyphens]{url}  
\usepackage{graphicx} 
\usepackage{natbib}  
\usepackage{caption} 
\usepackage{algorithm}
\usepackage{algorithmic}

\usepackage{newfloat}
\usepackage{listings}
\DeclareCaptionStyle{ruled}{labelfont=normalfont,labelsep=colon,strut=off} 
\floatstyle{ruled}
\newfloat{listing}{tb}{lst}{}
\floatname{listing}{Listing}
\usepackage[accsupp]{axessibility} 

\usepackage{amsmath}
\usepackage{amssymb}
\usepackage{booktabs}

\usepackage{color}

\usepackage{mathtools}
\usepackage{iac_pkg}
\usepackage{lipsum}
\usepackage{amssymb}
\usepackage{pifont}
\usepackage{multirow}
\usepackage{tabularx, booktabs}
\usepackage[htt]{hyphenat}
\usepackage{color}
\usepackage{xspace}
\usepackage{cite}
\usepackage{overpic}
\usepackage{arydshln}
\usepackage{subcaption}
\usepackage{xcolor}
\usepackage{bm}

\definecolor{citecolor}{RGB}{34,139,34}
\definecolor{lightred}{RGB}{255,100,100}
\definecolor{cell_bisque}{rgb}{1.0, 0.89, 0.77}
\definecolor{cell_blond}{rgb}{0.98, 0.94, 0.75}
\definecolor{cell_blue}{RGB}{155, 187, 228}
\definecolor{princetonorange}{rgb}{1.0, 0.56, 0.0}
\definecolor{pinkpearl}{rgb}{0.91, 0.67, 0.81}
\definecolor{mossgreen}{rgb}{0.68, 0.87, 0.68}

\newcommand{\Paragraph}[1]{\vspace{2mm}\noindent\textbf{#1.}\hspace{0mm}}
\newcommand{\Section}[1]{\vspace{0mm} \section{#1} \vspace{0mm}}
\newcommand{\SubSection}[1]{\vspace{0mm} \subsection{#1} \vspace{0mm}}

\title{Low-Light Image Enhancement via Generative Perceptual Priors}
\author{
    Han Zhou\textsuperscript{\rm 1}\equalcontrib, Wei Dong\textsuperscript{\rm 1}\equalcontrib, Xiaohong Liu\textsuperscript{\rm 2}\thanks{Corresponding Author}, Yulun Zhang\textsuperscript{\rm 2}, Guangtao Zhai\textsuperscript{\rm 2}, Jun Chen\textsuperscript{\rm 1}\\
}
\affiliations{
    \textsuperscript{\rm 1}McMaster University \quad \quad \textsuperscript{\rm 2}Shanghai Jiao Tong University\\ 
    \{zhouh115, dongw22, chenjun\}@mcmaster.ca \quad
    \{xiaohongliu, yulzhang, zhaiguangtao\}@sjtu.edu.cn\\

}

\begin{document}
\maketitle

\begin{figure*}  

\setlength{\abovecaptionskip}{0.7mm}
    \vspace{-10mm}
    \includegraphics[width = 0.96\textwidth]{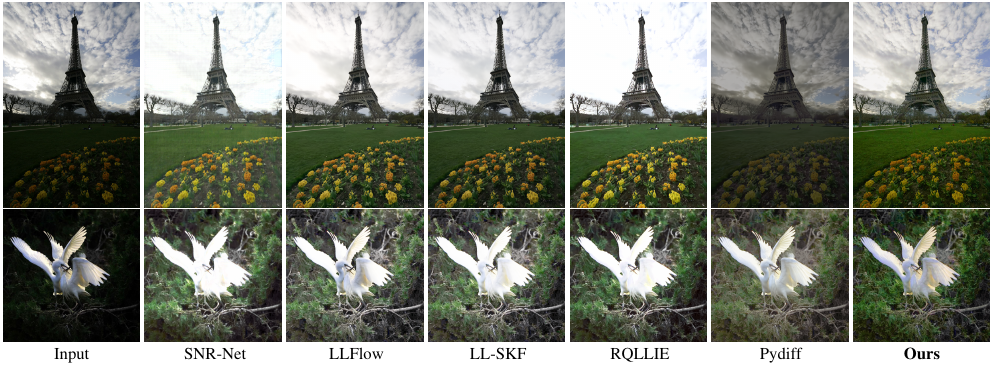}
    \captionof{figure}{Visual comparisons on real-world datasets without Ground Truth. The images analyzed are sourced from MEF~\cite{MEF} and NPE~\cite{NPE}, respectively. Our method stands out by providing a \textbf{balanced enhancement effect}. Unlike other approaches, our method effectively enhances luminance, not only to reveal finer details but also preserve natural color tones. Notably, the \textit{\textbf{cloud details}} in the sky, the \textit{\textbf{structural integrity}} of the Eiffel Tower and the  \textit{\textbf{texture of flowers and branches}} are rendered with clarity and \textit{\textbf{without over-exposure artifacts}} or unnatural coloration, distinctly surpassing others.}
    \vspace{1mm}
    \label{fig1}
\end{figure*}
\input{0_Sections/0_abstract}
\input{0_Sections/1_intro}

\input{0_Sections/2_rela}
\input{0_Sections/3_methods}

\input{0_Sections/4_experiments}
\input{0_Sections/5_experiments_ablation}

\input{0_Sections/6_conclusions}

\bibliography{main}
\end{document}

%% file: 0_Sections/0_abstract.tex
\begin{abstract}
Although significant progress has been made in enhancing visibility, retrieving texture details, and mitigating noise in Low-Light (LL) images, the challenge persists in applying current Low-Light Image Enhancement (LLIE) methods to real-world scenarios, primarily due to the diverse illumination conditions encountered. Furthermore, the quest for generating enhancements that are visually realistic and attractive remains an underexplored realm. In response to these challenges, we introduce a novel \textbf{LLIE} framework with the guidance of \textbf{G}enerative \textbf{P}erceptual \textbf{P}riors (\textbf{GPP-LLIE}) derived from vision-language models (VLMs). Specifically, we first propose a pipeline that guides VLMs to assess multiple visual attributes of the LL image and quantify the assessment to output the global and local perceptual priors. Subsequently, to incorporate these generative perceptual priors to benefit LLIE, we introduce a transformer-based backbone in the diffusion process, and develop a new layer normalization (\textit{\textbf{GPP-LN}}) and an attention mechanism (\textit{\textbf{LPP-Attn}}) guided by global and local perceptual priors. Extensive experiments demonstrate that our model outperforms current SOTA methods on paired LL datasets and exhibits superior generalization on real-world data. The code is released at \url{https://github.com/LowLevelAI/GPP-LLIE}.
\end{abstract}

%% file: 0_Sections/1_intro.tex
\section{Introduction}
\label{sec:intro}

Images captured in low-light conditions exhibit compromised quality, characterized by diminished visibility, reduced contrast, and loss of detail, which complicates both human observation and vision-based tasks~\cite{tracking2016, segmentation, segmentation2}. Enhancing these images is particularly challenging in real-world scenarios due to complex degradation caused by diverse illumination levels and elevated noise intensities, among others.

\input{1_Figures/Figure_LLaVA}

Although substantial progress has been made for low-light image enhancement (LLIE)~\cite{glare, FastLLVE, ecmamba}, traditional methods, which rely on handcrafted descriptors\cite{fu2016variationalmodel, Jobsob1997multiretinex, zheng2022semanticzeroshot} and existing deep learning approaches, which exploit the data fitting capability of neural networks~\cite{photoadjust2020acm, drbn2020cvpr,llnet2017}, still suffer from poor generalizability and adaptability in real-world scenarios. Image priors (\textit{i.e.}, edge)~\cite{smg2023cvpr, edgeguide2021, edgeiccv2021, edgeaaai2020}, semantic priors~\cite{semantic2023cvpr, issr2020acmmm}, and illumination maps~\cite{illumin2020cvpr} are commonly used for improving image quality. However, their effectiveness in generating realistic details is limited due to the difficulty of predicting accurate and robust priors from severely degraded inputs. Semantic priors are also constrained by the reliance on predefined semantic categories, which affects their ability to generalize in practical applications. Recent work~\cite{daclip} has attempted to fine-tune the pre-trained VLMs to align the content embedding for degraded images to  that for clean images. Yet, fine-tuning VLMs on limited data has a risk of over-fitting and demonstrates limited generalizability on unseen data.

On the other hand, several recent generative LLIE approaches~\cite{DiffLL, pydiff, retidiff, clediff} are proposed based on Diffusion Model (DM)~\cite{diffusion, diffusionbeatgan}, which demonstrates impressive capacity to represent intricate distributions without succumbing to the mode collapse and training instability often associated with GANs. Despite the prevailing choice of convolutional U-Net network as the de facto backbone, \cite{diffusiontransformer} indicates that the inductive biases inherent in U-Net may not be pivotal for the remarkable performance of diffusion models and introduces the transformer-based diffusion model (DiT) with good scalability for image synthesis. However, directly employing DiT for LLIE is unfeasible: \textbf{(1)} The original DiT network is designed to output images with specific resolutions, whereas LLIE models typically process images of varying sizes; \textbf{(2)} The computational complexity of the Vision Transformer in DiT scales quadratically with the input size, which limits its applicability to high-resolution images; \textbf{(3)} Lacking prior information, the original DiT often demonstrates limited generalization to real-world images.

In view of the aforementioned challenges, we propose a novel \textbf{LLIE} framework with the guidance of \textbf{G}enerative \textbf{P}erceptual \textbf{P}riors (\textbf{GPP-LLIE}) derived from vision-language models (VLMs). Firstly, we propose incorporating external perceptual priors to help LLIE model to comprehend the varying attributes in LL images, thereby guiding the enhancement process. Different from previous strategies that extract features from severely degraded images, we develop a novel pipeline as shown in Fig.~\ref{fig_llava} to extract generative perceptual priors from VLMs, which are pre-trained on extensive datasets and exhibit proficiency in perceiving visual low-level attributes on unseen images. Specifically, we develop instruction prompts to guide VLMs to assess multiple attributes (\textit{i.e.}, contrast, visibility, and sharpness) of LL images globally and locally. Then, these assessments are quantified into the global and local perceptual priors via our proposed sigmoid-based quantification strategy. Secondly, we develop an efficient transformer-based backbone with the guidance of generative perceptual priors in the diffusion process. Specifically, the global perceptual prior is employed to modulate the layer normalization (\textit{\textbf{GPP-LN}}) and the local perceptual prior is introduced to guide the attention mechanism (\textit{\textbf{LPP-Attn}}). Experimental results attest to the effectiveness of our proposed network across diverse real-world scenarios, yielding visually appealing and detail-rich enhancement results as shown in Fig.~\ref{fig1} In addition, we also apply our generative perceptual priors to other LLIE models and enhanced results further demonstrate the effectiveness of our proposed pipeline of extracting generative perceptual priors from VLMs. We highlight our contributions as follows:

(1) We introduce an \textbf{innovative pipeline} to acquire \textbf{generative perceptual priors} for LL images globally and locally based on pre-trained VLMs.

(2) With the guidance of global and local generative perceptual priors, we develop an efficient transformer based diffusion framework for LLIE (\textbf{GPP-LLIE}).

(3) We introduce global perceptual priors to modulate the layer normalization (\textbf{GPP-LN}) and leverage local perceptual priors to guide the attention mechanism (\textbf{LPP-Attn}) in our transformer block to benefit the enhancement process.

(4) Our method demonstrates SOTA performance on \textbf{various benchmark datasets} and exhibits \textbf{good generalization on real-world data}. Furthermore, our generative perceptual priors are applicable to help other LLIE models achieve enhanced outcomes.

%% file: 1_Figures/Figure_LLaVA.tex
\begin{figure*}[t]
    \centering
    \setlength{\abovecaptionskip}{1mm}
    \centering
    \includegraphics[width=0.94\linewidth]{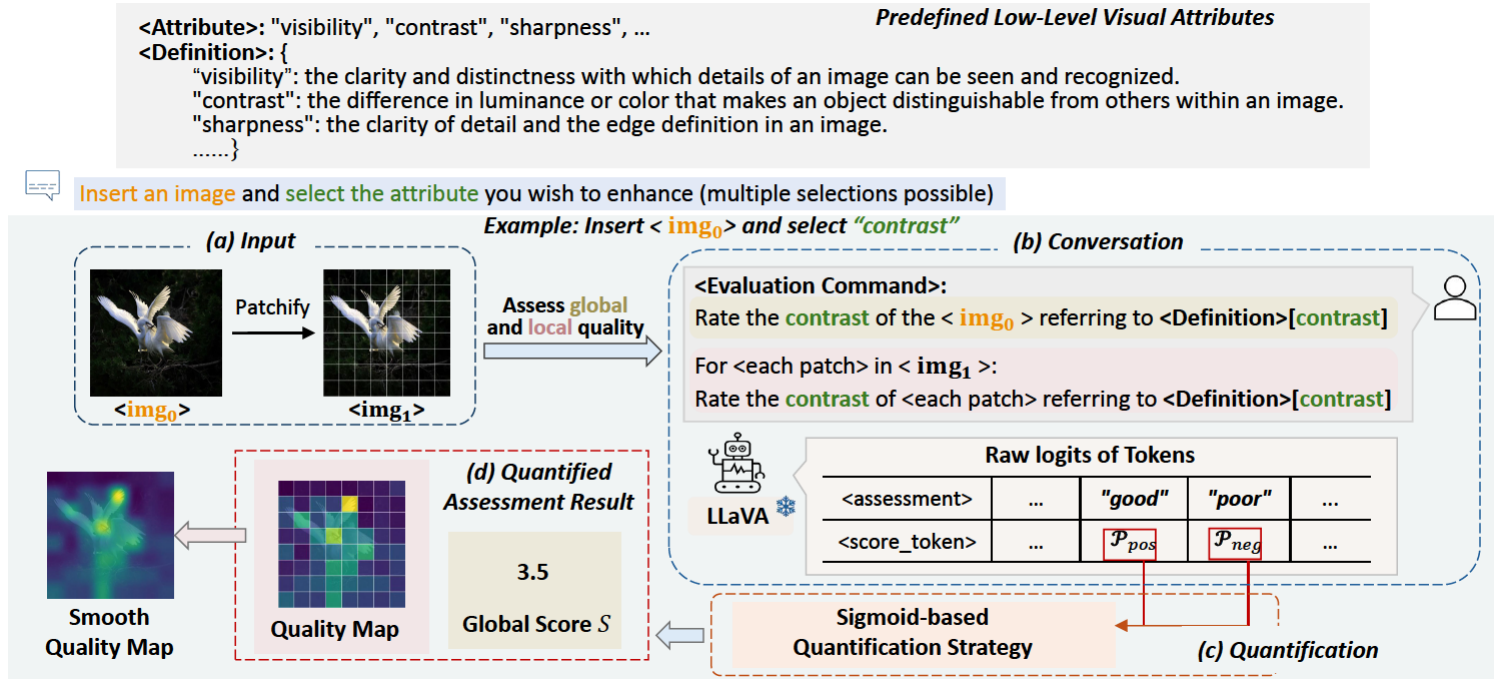}
    \caption{Our proposed pipeline for fine-grained \textbf{generative perceptual priors extraction} from pre-trained VLMs. (a) The input image is patchified into several non-overlapped patches. (b) Conversation with VLMs. We develop evaluation commands to guide VLMs to assess the image globally and locally regarding the selected attribute. (c) Based on the output of VLMs, we design the sigmoid-based quantification strategy. (d) Our extracted global and local generative priors.}
    \label{fig_llava}
    \vspace{-3.9mm}
\end{figure*}

%% file: 0_Sections/2_rela.tex
\section{Related Works}
\label{sec:rela}

\paragraph{Diffusion-Based LLIE} Compared to common deep learning architectures for image restoration~\cite{GridDehazeNet, GridDehazeNet+, dehazedct2024cvprw, breakhaze2023han, shadowremoval2024cvprw, NTIRE_Dehazing_2023, NTIRE_Dehazing_2024, vasluianu2024ntire_isr, FMSNet, AttentionLut},
due to its remarkable generative capabilities, diffusion model~\cite{diffusion, diffusionbeatgan} has emerged as a dominant force in recently proposed image-level generative models, finding widespread applications in image synthesis~\cite{diffusiontransformer}, inpainting~\cite{ldm}, and super-resolution~\cite{coser}. Diffusion model is firstly introduced for LLIE in Diff-Retinex~\cite{diff-retinex} which employs two diffusion networks to reconstruct normal-light illumination and reflectance maps to produce the final result. Then, Reti-Diff~\cite{retidiff} proposes to utilize the diffusion process in the latent space to alleviate computational costs and leverage a transformer network for detail refinement. CLEDiff~\cite{clediff} introduces a diffusion network conditioned on brightness level. PyDiff~\cite{pydiff} adopts a strategy that progressively increases the resolution in the reverse diffusion process. However, all these methods utilize convolutional U-Nets architecture as the backbone for diffusion model. The untapped potential of Transformers, renowned for their scalability and effectiveness, remains unexplored in diffusion-based LLIE methods.

\paragraph{Image Restoration with Vision-Language Guidance}
Pre-trained with large-scale image-text datasets, vision-language models (\textit{i.e.}, CLIP~\cite{CLIP}) demonstrate impressive zero-shot capability and have been applied to image restoration in recent years. For example, CLIP-LIT~\cite{clip_lit} introduces learnable prompts for unsupervised backlit image enhancement. DACLIP~~\cite{daclip} designs an additional image controller to separate degradation and image features from the frozen CLIP latent space for universal image restoration. These methods mainly concentrate on applying VLMs to generate captions of image content. However, accurately describing the content of low-light, complex scene images presents a challenge. Instead, based on pre-trained VLMs, we propose an innovative pipeline to extract generative perceptual priors regarding the low-level attributes of LL images. Moreover, we incorporate these priors into our proposed transformer-based diffusion network to enhance the performance in LLIE. 

%% file: 0_Sections/3_methods.tex
\section{Method}
\label{sec:method}
\input{1_Figures/Figure_framework}

The main focus of this work is to extract generative perceptual priors that well represent visual attributes of LL images and to develop LLIE models guided by these priors to generate realistic and visually attractive enhancement results. The overall framework is illustrated in Fig.~\ref{fig_framework}.

In this section, we first discuss the motivation of employing guidance derived from vision language models (VLMs) for LLIE task (Sec.~\ref{sec_method_motivation}). Then, we propose an innovative pipeline that guides VLMs to assess the visual attributes of LL images globally and locally and then extract the perceptual priors by introducing the sigmoid-based quantification strategy (Sec.~\ref{sec_llm}). Moreover, we develop a transformer based diffusion framework and we incorporate these priors to guide the reverse diffusion process (Sec.~\ref{sec_trandformer_block}).

\SubSection{Motivation of utilizing VLMs Guidance} 
\label{sec_method_motivation}

Although recent methods in low-light image enhancement (LLIE) have shown improved performance, they often yield unbalanced results with over-exposure artifacts when applied to real-world images, which frequently differ in lighting conditions from the training datasets (see Fig.~\ref{fig1}). These results underscore a general inability of current LLIE methods to adaptively enhance images under varied illumination conditions. Thus, enabling models to autonomously \textit{\textbf{perceive}} and adapt to various visual distortions is of vital importance. Inspired by the recently demonstrated capabilities of emergent Vision-Language Models (VLMs) in low-level visual perception and understanding~\cite{Q-Instruct}, we aim to explore the potential of utilizing these perceptual abilities of VLMs to facilitate LLIE tasks.

\SubSection{Generative Perceptual Priors from VLMs}
\label{sec_llm}
VLMs are usually trained with millions of text-image pairs and demonstrate remarkable zero-shot capabilities in generating aligned understandings between texts and images. Therefore, it is essentially promising to utilize the prior information inherent in VLMs to help LLIE models make more appropriate decisions during the restoration process. However, the VLMs employed in recent image restoration works~\cite{daclip, coser} are primarily focused on understanding the semantic content of images, yet they lack precise representation of visual details. Moreover, accurately describing the content of complex LL images is quite challenging. In contrast, the VLMs employed in our work is LLaVA \cite{LLaVA}, which is further fine-tuned with 200K instruction-response pairs related to low-level visual aspects in Q-instruct \cite{Q-Instruct}. In this paper, we introduce a new pipeline to employ LLaVA in LLIE: we design text prompts to guide LLaVA to assess multiple visual attributes of LL images. In addition, different from the text/image embedding in previous methods \cite{coser, daclip}, we introduce the quantification strategy to output the quantified global assessment and local quality map as perceptual priors for LLIE. Our pipeline for perceptual priors extraction is shown as Fig.~\ref{fig_llava}.

\Paragraph{Fine-grained Low-Level Visual Assessment} Different from simply evaluating the overall quality of image, we offer several low-level visual attributes \textbf{\texttt{<}Attribute\texttt{>}} for selection and provide corresponding definitions \textbf{\texttt{<}Definition\texttt{>}} to help VLMs better understand the assessment task. Specifically, given an image \textbf{\texttt{<}img$\mathbf{_0}$\texttt{>}}, we can select the attribute from \textbf{\texttt{<}Attribute\texttt{>}} for assessment. For example, as LLIE is to enhance the contrast, visibility, and sharpness of LL images, we can sequentially evaluate these attributes in LL inputs. 

Besides the global assessment, considering the varying contrast, visibility, and sharpness within LL images, we also propose to extract local assessment for fine-grained enhancement. Specifically, the input \textbf{\texttt{<}img$\mathbf{_0}$\texttt{>}} is patchified into several non-overlapped patches \textbf{\texttt{<}img$\mathbf{_1}$\texttt{>}} and each patch is fed into LLaVA to acquire the local assessment. Therefore, the overall evaluation command during the conversation is defined as the \textbf{\texttt{<}Evaluation Command\texttt{>}} in Fig.~\ref{fig_llava}.

\Paragraph{Sigmoid-based Quantification Strategy} With the input image and evaluation command, we observe several issues in the probabilities of tokens generated by LLaVA: (1) Tokens (\textit{i.e.}, \textbf{\textit{``The''}}) that have the highest probability are meaningless. (2) Compared to a single token, the difference between two opposite logits (\textit{i.e.}, \textbf{\textit{``good''}} and \textbf{\textit{``poor''}}) is more aligned with human perception. These observations motivate us to explore additional strategies to generate quantified output well-aligned with human opinions for LLIE.

To start with, we design our strategy based on tokens with contextual information rather than those with the highest probability. Moreover, our strategy is designed based on the probability of \textbf{\textit{``good''}} and \textbf{\textit{``poor''}}, which can be considered as positive and negative assessments of the attributes of LL images. This strategy aligns more closely with human perceptual system, as assessments of images typically include both positive and negative evaluations. Therefore, the quantified global score $S$ is calculated as $S=(1+e^{-(\mathcal{P}_{pos}-\mathcal{P}_{neg})/\alpha})^{-1}$, where a sigmoid operation modulates the difference between the probabilities of \textbf{\textit{``good"}} and \textbf{\textit{``poor"}} (denoted as $\mathcal{P}_{pos}$ and $\mathcal{P}_{neg}$), $\alpha$ is the modulation scalar and is set as $3$ in this work. Similarly, we calculate the score for each patch in \textbf{\texttt{<}img$\mathbf{_1}$\texttt{>}} and then obtain the quality map \textbf{\textit{M}}. Specific to our LLIE task, three attributes (``contrast'', ``visibility'', and ``sharpness'') are evaluated and the average global score and concatenated quality map are introduced as perceptual prior guidance in our proposed LLIE model.

\input{2_Tables/table1}

\SubSection{Generative Perceptual Prior Guided Diffusion Transformer} 
\label{sec_trandformer_block}

To achieve enhanced generalizability on unseen real-world images, we build our LLIE model based on the Diffusion Transformer (DiT) network~\cite{diffusiontransformer}, which shares similar architecture with Vision Transformers (ViT) and presents good scalability properties. However, the DiT is originally designed for image synthesis at specific resolutions (\textit{i.e.}, $256 \times 256$ or $512 \times 512$) and the computational complexity of ViT is quadratic to the input size. Evidently, the original DiT is infeasible for LLIE task, as LLIE models usually process LL images with variable sizes and sometimes large resolutions. To this end, we introduce a transformer-based backbone in the diffusion process, which is suitable for LLIE and contains special designs for incorporating external generative perceptual priors.

\Paragraph{Overview} The overall framework of our generative perceptual priors guided diffusion transformer is shown as Fig.~\ref{fig_framework}. Given a paired normal-light (NL) image $\mbf{I}_{nl} \in \mathbb{R}^{H \PLH W \PLH 3 }$ and low-light (LL) image $\mbf{I}_{ll} \in \mathbb{R}^{H \PLH W \PLH 3 }$, an encoder $\mathcal{E}$ is employed to extract their latent representations $\mbf{z}^{0}_{nl} \in \mathbb{R}^{\frac{H}{f} \PLH \frac{W}{f} \PLH d }$ and $\mbf{z}_{ll} \in \mathbb{R}^{\frac{H}{f} \PLH \frac{W}{f} \PLH d }$, where $H$, $W$, $d$, and $f$ represent the image height, image width, the hidden dimension, and the down-sampling factor of $\mathcal{E}$. Then, the forward diffusion process is applied upon $\mbf{z}^{0}_{nl}$ and its noised version is denoted as $\mbf{z}^{T}_{nl}$. For the reverse demonising process, we gradually transform the randomly sampled Gaussian noise $\mbf{\hat{z}}^{T}_{nl}$ into a clear NL latent feature $\mbf{\hat{z}}^{0}_{nl}$ step by step. For each step $t$, besides the latent representation for LL image $\mbf{z}_{ll}$, we also incorporate the generative perceptual prior extracted from LLaVA (shown in Fig.\ref{fig_llava}) as the guidance in our proposed GPP-LLIE network $\mathbb{\epsilon_\theta}$. Finally, the restored feature $\mbf{\hat{z}}^{0}_{nl}$ is fed into the decoder $\mathcal{D}$ to produce the final result $\mbf{\hat{I}}_{out}$.
 
\input{1_Figures/Fig_ALL_lol_4}

\Paragraph{GPP-LLIE Network} Our GPP-LLIE network $\mathbb{\epsilon_\theta}$ is shown as Fig.~\ref{fig_framework}, highlighting its several unique characteristics:

\textit{\textbf{Concat-and-Remove Strategy}}: Within each GPP-LLIE block, we first concatenate the LL feature $\mbf{z}_{ll}$ to the input to introduce the LL information into the reverse diffusion process to enhance the fidelity. While at the end of the block, we remove the latter half of the channels, enabling the concatenation of the $\mbf{z}_{ll}$ at the start of the next GPP-LLIE block.

\textit{\textbf{Global Perceptual Prior Guided Layer Norm (GPP-LN)}}: To effectively integrate the global score $S$ derived from VLMs into our GPP-LLIE block, we modulate the layer normalization process. This modulation, driven by the scale and shift parameters ($\gamma$ and $\beta$) influenced by $S$, optimizes the normalization process to better reflect the perceptual insights provided by the global perceptual prior. Given an input feature $\mbf{z}_{in}$, the output of our GPP-LN operation is calculated by: $\mbf{z}_{out} = \gamma \cdot {\rm LN} (\mbf{z}_{in}) + \beta$, where $\gamma, \beta = {\rm MLP}(S)$.

\textit{\textbf{Local Perceptual Prior Guided Attention (LPP-Attn)}}: To reduce the huge computational cost caused by spatial self-attention mechanism, we calculate the attention map along the channel dimension in our GPP-LLIE block. Moreover, beside the MSA, we also develop another channel attention mechanism guided by the local quality map \textbf{\textit{M}}. Specifically, \textit{query} element is calculated upon the input feature, while the calculation of \textit{key} and \textit{value} elements are guided by the local perceptual prior \textbf{\textit{M}}. 
Moreover, to facilitate the application of our LLIE model to LL images of diverse sizes, we remove the positional embedding from the Vision Transformer. Instead, the spatial positional embedding are learned with the guidance of the local perceptual prior \textbf{\textit{M}}.

%% file: 1_Figures/Figure_framework.tex
\begin{figure*}[!t]
    \centering
    \setlength{\abovecaptionskip}{1mm}
    \centering
    \includegraphics[width=\linewidth]{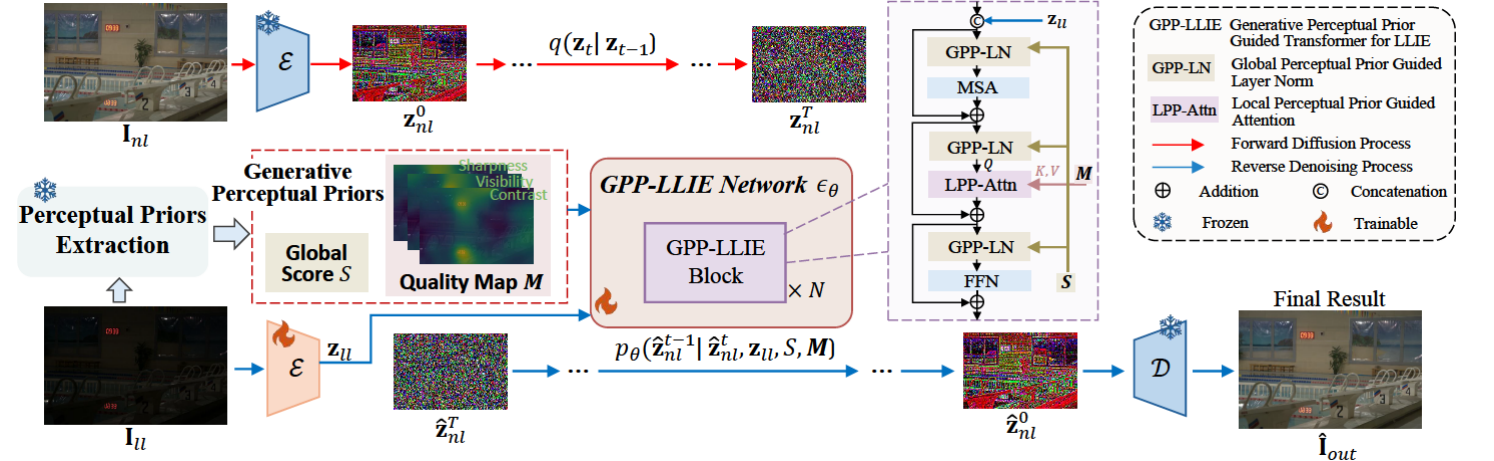}
    \caption{The overall framework of our proposed \textbf{GPP-LLIE} method. We first employ the encoder $\mathcal{E}$ to convert the NL image $\mbf{I}_{nl}$ and LL image $\mbf{I}_{ll}$ into latent space denoted as $\mbf{z}^{0}_{nl}$ and $\mbf{z}_{ll}$. Then, the forward diffusion process is applied upon $\mbf{z}^{0}_{nl}$. In order to leverage the prior information to guide the reverse diffusion process, the $\mbf{I}_{ll}$ is sent to our proposed pipeline for perceptual priors extraction. With the guidance of global perceptual score $S$, local quality map \textbf{\textit{M}}, and LL feature $\mbf{z}_{ll}$, we develop a transformer-based network $\epsilon_{\theta}$ and gradually transform the randomly sampled Gaussian noise $\mbf{\hat{z}}^{T}_{nl}$ into a clear NL latent feature $\mbf{\hat{z}}^{0}_{nl}$. Finally, the restored feature $\mbf{\hat{z}}^{0}_{nl}$ is fed into the decoder $\mathcal{D}$ to generate the final enhancement $\mbf{\hat{I}}_{out}$. }
    \label{fig_framework}
\end{figure*}

%% file: 2_Tables/table1.tex
\begin{table*}[t]
\Large
\setlength{\abovecaptionskip}{2mm}
\renewcommand\arraystretch{1}
  \centering
  \resizebox{\linewidth}{!}{
  \begin{tabular}{c|cccc|cccc|cccc}
    \hline
    \multirow{2}{*}{\begin{tabular}{c}
            \textbf{ Methods}
        \end{tabular}}&\multicolumn{4}{c|}{LOL} &\multicolumn{4}{c|}{LOL-v$2$-real}&\multicolumn{4}{c}{LOL-v$2$-syn}\\\cline{2-13}
        &FID $\downarrow$ &LPIPS$\downarrow$ &DISTS$\downarrow$ &PSNR$\uparrow$ &FID $\downarrow$ &LPIPS$\downarrow$ &DISTS$\downarrow$ &PSNR$\uparrow$  &FID $\downarrow$ &LPIPS$\downarrow$ &DISTS$\downarrow$ &PSNR$\uparrow$\\ 
   \hline
    KinD~\cite{KinD} &$78.28 $ &$0.157$ &$0.110 $ &$19.03$ &$95.02 $ &$0.151$ &$0.112 $ &$18.05$ &$97.32$ &$0.263$ &$0.180$ &$16.81$ \\
    MIRNet~\cite{MIRNet} &$71.16$ &$0.131 $ &$0.105$&$24.14$ &$82.25$ &$0.138$ &$0.116$&$20.02$ &$40.18$ &$0.102$ &$0.126$ &$21.94$\\
    DRBN~\cite{drbn2020cvpr}&$85.57$ &$0.155$ &$0.108$ &$19.86$ &$94.22$ &$0.147$ &$0.119$ &$20.29$ &$28.74 $ &$0.085$ &$0.097$ &$23.22$\\
    SNR-Net~\cite{SNR}&$66.47$ &$0.115$ &$0.092$ &$24.70$  &$68.56 $  &$0.120$ &$0.095$ &$21.48$ &$19.96$ &$0.056$ &$0.063$ &$24.14$\\
    URetinex-Net~\cite{URetinex-Net} &$85.59$ &$0.121$ &$0.096$ &$21.33$  &$76.74$ &$0.144$&$0.107$ &$21.16$ &$33.25$ &$0.075$ &$0.087$ &$24.73$\\   
    LLFlow~\cite{llflow2022}&$65.17$ &$0.113$ &$0.094$ &$25.19$ &$70.68 $ &$0.135$  &$0.102$ &$26.53$ &$20.24$ &$0.044$  &$0.056$ &$26.23$ \\
    
    RQLLIE~\cite{vqiccv23}  &$53.32$ &$0.121$ &$0.086$ &$25.24$  &$68.89 $ &$0.142 $ &$0.102$ &$22.37$ &$16.96$ &$0.044$ &$0.053$ &$25.94$  \\
    Retinexformer~\cite{retinexformer}&$72.38$ &$0.131$ &$0.106$ &$25.16$ &$79.58$ &$0.171$ &$0.115$ &$22.80$  &$22.78$ &$0.059$ &$0.066$ &$25.67$\\
    CUE~\cite{cueiccv23}  &$69.83$ &$0.224$ &$0.141$ &$21.86$  &$67.05$ &$0.133$ &$0.112$ &$21.19$ &$31.33$ &$0.076$ &$0.083$ &$24.41$  \\
    CLEDiff~\cite{clediff}  &$86.94$ &$0.164$ &$0.108$ &$25.51$  &$82.27$  &$0.183$ &$0.118$ &$22.68$ &$18.58$ &$0.064$ &$0.080$ &$27.38$  \\
    LL-SKF~\cite{semantic2023cvpr}&$59.47$ &$0.105$ &$0.084$ &$26.80$  &$57.84$ &$0.111$ &$0.084$ &$28.45$ &$21.58$ &$0.040$  &$0.063$ &$29.11$\\
    Reti-Diff~\cite{retidiff}  &$49.14$ &$0.105$ &$0.082$ &$25.35$ &$\blue{43.18}$ &$\blue{0.087}$ &$0.069$ &$22.97$ &$\blue{13.26}$ &$\blue{0.038}$ &$0.051$ &$27.53$ \\
    PyDiff~\cite{pydiff}  &$48.28$ &$\blue{0.099}$ &$\blue{0.079}$ &$\blue{27.09}$  &$44.27$ &$0.094$ & $0.072$ &$\blue{28.77}$ &$15.69$ &$0.040$ &$0.055$ &$\blue{29.27}$ \\
    DiffLL~\cite{wcdm}  &$\blue{48.11}$ &$0.118$ &$0.091$ &$26.33$  &$45.36$ &$0.089$ &$\blue{0.064}$ &$28.66$ &$13.66$ &$\blue{0.038}$ &$\blue{0.047}$ &$28.87$  \\

    \hline
    
    \textbf{Ours}  &{\red{$\mathbf{36.73}$}} &{\red{$\mathbf{0.081}$}} &{\red{$\mathbf{0.063}$}} &{\red{$\mathbf{27.51}$}}  &{\red{$\mathbf{26.78}$}} &{\red{$\mathbf{0.055}$}} &{\red{$\mathbf{0.047}$}} &{\red{$\mathbf{29.23}$}} &{\red{$\mathbf{9.74}$}} &{\red{$\mathbf{0.031}$}} &{\red{$\mathbf{0.039}$}} &{\red{$\mathbf{30.17}$}}  \\
  \hline
  \end{tabular}
  }
  \caption{Quantitative comparisons on LOL, LOL-v2-real, and LOL-v$2$-synthetic, our method achieves superior performance compared to current SOTA methods. These numbers are obtained either by using their released weights or by re-training their models. [Key: \textbf{\red{Best}}, \blue{Second Best}, $\uparrow$ ($\downarrow$): Larger (smaller) values leads to better performance] \vspace{-3mm}}
  \label{table_1}
\end{table*}

%% file: 1_Figures/Fig_ALL_lol_4.tex
\begin{figure*}[!t]
    \centering
    \setlength{\abovecaptionskip}{1mm}
    \centering
    \includegraphics[width=0.97\linewidth]{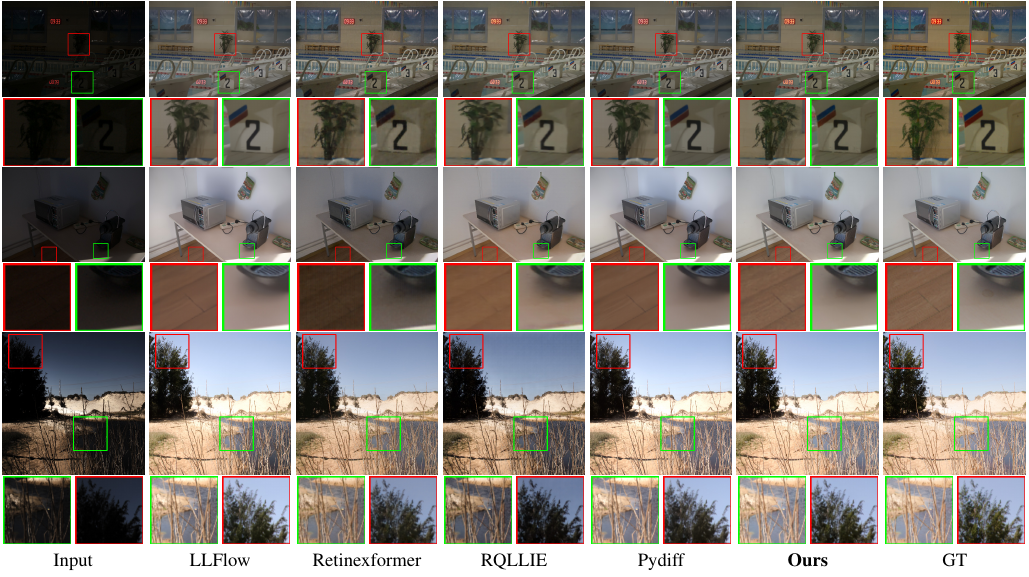}
    \caption{Visual comparisons on paired dataset. The images are sourced respectively from LOL (row $1$), LOL-v2-real (row $2$), and LOL-v2-syn (row$3$). Previous methods often result in overly smoothed images, consequently obscuring pivotal textural details. In contrast, our method yields sharper images while retaining the delicate details. For instance, our method maintains the structural complexities of the foliage and branches within the potted plant (first row). In second row, the grain on the wooden floor surface, as well as its edge contours, have been well-preserved. Similarly, in the natural landscapes, our method excels at enhancing the clarity of twig edges and maintaining color consistency.}
    \label{fig:LOL}
\end{figure*}

%% file: 0_Sections/4_experiments.tex
\Section{Experiments}
\label{sec_ex}

\SubSection{Experiment Settings} \label{sec_imple}
\Paragraph{Training and Diffusion} Our model is trained using the AdamW optimizer with the total training iterations of $1.5$M for all datasets and the learning rate is set to $10^{-4}$. Each training input is cropped into $320 \times 320$, and the batch size is set to $16$. We use horizontal flips and rotations for data augmentation. For the diffusion process, the total timesteps for training is set to $1,000$ during the training, and we use $25$ steps to accelerate sampling process for the inference.

\Paragraph{Low Light Datasets and Metrics}
We conduct experiments on various low light datasets including LOL~\cite{LOLv1}, LOL-v2-real, and LOL-v2-synthetic~\cite{LOLv2}. Specifically, we train our model using 485, 689, and 900 LL-NL pairs on LOL, LOL-v2-real, and LOL-v2-synthetic datasets, and other 15, 100, and 100 images are used for evaluation. Moreover, we also test the generalization of our method on several real-world datasets without ground truth images including MEF~\cite{MEF}, LIME~\cite{LIME}, DICM~\cite{DICM}, and NPE~\cite{NPE}. Metrics for paired datasets includes Fr\'{e}chet Inception Distance (FID)~\cite{fid}, Learned Perceptual Image Patch Similarity (LPIPS)~\cite{LPIPS}, Deep Image Structure and Texture Similarity (DISTS)~\cite{dists}, and Peak Signal-to-Noise Ratio (PSNR). Besides, we adopt a no-reference metric Natural Image Quality Evaluator (NIQE)~\cite{NIQE} for unpaired real data evaluation.

\subsection{Performance on LLIE}
\label{sec_ex_LLIE}
\Paragraph {Quantitative Results on Paired Dataset} Tab.~\ref{table_1} summarizes the quantitative comparisons between our method with current SOTA methods. Our method achieves the superior performance on $3$ commonly-used benchmarks, highlighting its advance and perfect generalization. Notably, the FID scores of our method surpass the best of SOTA methods by $\bm{23.6}\%$, $\bm{37.4}\%$, and $\bm{26.5}\%$ on the LOL, LOL-v2-real, and LOL-v2-synthetic, respectively. Moreover, our LPIPS values significantly outperforms the PyDiff by $\bm{18.4}\%$. These numbers demonstrate the superior perceptual quality of our enhancement and prove the effectiveness of generative perceptual priors in our method. Besides, our leading DISTS and PSNR scores show the satisfactory ability of our model in recovering texture details and maintaining fidelity.

\Paragraph {Qualitative Results on Paired Datasets} We present the enhanced images of different methods in Fig.~\ref{fig:LOL}. Our appealing and realistic enhancement results demonstrate our method can generate images with pleasant illumination, correct color retrieval, and enhanced texture details. For example, the rich structural details of potted plants in row $1$, the well preserved wooden floor surface and its edge contours in row $2$, and the enhanced twig edge in row $3$. In contrast, previous methods tend to output blurry results without rich textures (Retinexformer and PyDiff in row $2$) and struggle to preserve color fidelity and illumination harmonization (LLFlow and RQ-LLIE in row $1$). 

\input{1_Figures/DICM_FIG5}
\input{2_Tables/table_unpair2}

\Paragraph {Performances on Real-world Dataset} To further verify the generalization of our method, we extend our evaluations to real-world low light datasets. Notably, we select several methods that perform well in Tab.~\ref{table_1} and we use the weights of various methods trained on LOL training split to ensure fairness. The quantitative comparisons are presented as Tab.~\ref{table_unpaired}, where the NIQE metric is employed. It can be seen that our model achieve the best performance on the first $3$ datasets and the second best result on NPE, thereby significantly outperforming other methods across real-world data. 

Moreover, visual comparisons are reported in Fig.~\ref{fig1} and Fig.~\ref{fig_visual_unpair1}, where previous methods tend to output overexposure results or struggle to preserve details. In contrast, our enhanced images are more realistic and appealing and our method can flexibly enhance images with various illuminations. Specifically, for outdoor images with uneven illumination distribution, our method enhances the brightness while simultaneously preventing local overexposure, thereby preserving more details (\textit{i.e.}, the enhanced architectural details with sharper edges and preserved details in row $1$ of Fig.~\ref{fig_visual_unpair1}, and the improved contrast in the sky and cloud area in row $2$ of Fig.~\ref{fig_visual_unpair1} . These vivid and natural enhanced images, together with quantitative numbers reported in Tab.~\ref{table_1} and Tab.~\ref{table_unpaired}, demonstrate the superior effectiveness and generalization of our method compared to the current SOTA. 

%% file: 1_Figures/DICM_FIG5.tex
\begin{figure*}[!t]
    \centering
    \setlength{\abovecaptionskip}{1mm}
    \centering
    \includegraphics[width=0.97\linewidth]{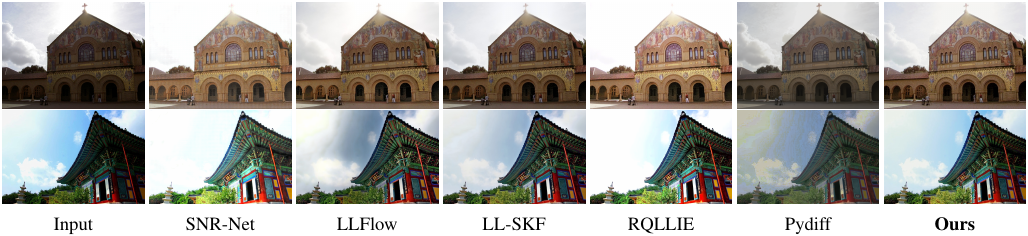}
    \caption{Visual comparisons on real-world datasets (DICM~\cite{DICM}). Our method adeptly handles the \textit{\textbf{diverse and uneven}} illumination levels present in the original image. It effectively enhances brightness and contrast while \textit{\textbf{avoiding the overexposure}} of original bright areas and \textit{\textbf{maintaining natural coloration}}, generating visually appealing results.}
   \label{fig_visual_unpair1}

\end{figure*}

%% file: 2_Tables/table_unpair2.tex
\begin{table}[t]
\renewcommand\arraystretch{1.25}
\setlength{\abovecaptionskip}{2mm}
  \centering
  \resizebox{\linewidth}{!}{
  \begin{tabular}{c|cccc|c}
    \hline
        
    \textbf{Methods} &MEF &LIME &DICM &NPE &Mean\\
    \hline 
    SNR-Net~\cite{SNR} &$4.14$ &$5.51$ &$4.62$ &$4.36$ &$4.60$ \\
    LLFlow~\cite{llflow2022}  &\blue{$3.92$} &$5.29$ &$3.78$ &$4.16$ &$3.98$ \\
    LL-SKF~\cite{semantic2023cvpr} &$4.03$ &$5.15$  &\blue{$3.70$} &$4.08$ &\blue{$3.92$} \\
    RQLLIE~\cite{vqiccv23} &$4.21$ &\blue{$4.86$} &$4.02$ &\red{$\bm{4.03}$} &$4.13$ \\
    CLEDiff~\cite{clediff} &$5.27$ &$5.00$ &$4.42$  &$4.57$ &$4.57$ \\
    PyDiff~\cite{pydiff} &$4.24$ &$4.88$ &$4.32$ &$4.38$ &$4.37$  \\
    \hline
    \bf{Ours} &\red{$\bm{3.55}$} &\red{$\bm{4.24}$} &\red{$\bm{3.58}$} &\blue{$4.05$} &\red{$\bm{3.67}$} \\
    \hline
    \end{tabular}
  }
  \caption{Quantitative comparisons on real-world datasets in terms of NIQE. These numbers are obtained by testing with their released LOL weights. [Key: \textbf{\red{Best}}, \blue{Second Best}]}
  \label{table_unpaired}
\end{table}

%% file: 0_Sections/5_experiments_ablation.tex
\SubSection{Ablation Study}
\label{sec_ex_abla}

\input{2_Tables/table_abla}

To analyze the contribution of each component in our method, we conduct extensive ablation studies.

\Paragraph {Local Perceptual Prior and LPP-Attn} To study the importance of local perceptual prior and our proposed LPP-Attn, we remove these two parts from our model and denote the remaining network as Variant 1. Tab.~\ref{tab_ab_local} reports the quantitative performance of Variant 1 on LOL dataset, which still presents competitive enhancement performance on all measured metrics compared to the current SOTA in Tab.~\ref{table_1}. However, compared to Variant 1, our full GPP-LLIE shows significantly enhanced FID and LPIPS scores by integrating the local perceptual prior using our proposed LPP-Attn mechanism. Besides, we introduce the Variant 2 that replaces our LPP-Attn with spatial feature transform layer~\cite{SFT} applied in StableSR~\cite{StableSR}, our LPP-Attn mechanism achieves 22\%, 19\%, and 22\% lower values in FID, LPIPS, and DISTS respectively. These comparisons manifest the importance of our propose LPP-Attn and the local perceptual prior generated by our proposed extraction pipeline in Fig.~\ref{fig_llava}.

\Paragraph {Global Perceptual Prior and GPP-LN} Based on Variant 1, we implement several adaptations to illustrate the effectiveness of our extracted global perceptual prior and the corresponding GPP-LN. (1) We remove global perceptual prior and GPP-LN and denote the remaining network as Variant 3. (2) We remove GPP-LN operation and directly add the global score to the noised latent feature $\mbf{\hat{z}}^{T}_{nl}$. This framework is referenced as Variant 4. The quantitative results of Variant 3 and Variant 4 are reported in Tab.~\ref{tab_ab_global}. The discernible performance disparity between Variant 3 and Variant 1 (\textit{i.e.}, 23\% higher FID and 10\% higher in LPIPS) illustrates the superiority of the global perceptual score and our proposed GPP-LN. Besides, compared to Variant 4, the GPP-LN employed in Variant 1 better incorporates the prior information and guides our model achieve better performance (\textit{i.e.}, 14.6\% lower in FID and 7.2\% lower in LPIPS).

%% file: 2_Tables/table_abla.tex
\begin{table}[t]
\renewcommand\arraystretch{1.25}
\setlength{\abovecaptionskip}{1mm}
\centering
    \begin{subtable}[t]{1\linewidth}
    \centering
    \resizebox{0.8\linewidth}{!}{
    \begin{tabular}{c|cccc}
    \hline
     Model &FID $\downarrow$ &LPIPS$\downarrow$ &DISTS$\downarrow$ &PSNR$\uparrow$  \\ 
    \hline
    \textbf{Ours} &\red{$\bm{36.73}$} &\red{$\bm{0.081}$} &\red{$\bm{0.063}$} &\red{$\bm{27.51}$} \\
    Variant 1 &$49.83$ &$0.103$ &$0.084$ &$26.88$ \\
    Variant 2 &\blue{$47.18$} &\blue{$0.100$} &\blue{$0.081$} &\blue{$27.06$} \\
    \hline
    \end{tabular}
    }
    \caption{By integrating the local perceptual prior using our proposed LPP-Attn mechanism, GPP-LLIE shows enhanced performance in terms of FID and LPIPS. Besides, our LPP-Attn performs better than spatial feature transform applied in StableSR.}
    \label{tab_ab_local}
    \end{subtable}
    
    \begin{subtable}[t]{1\linewidth}
    \centering 
    \resizebox{0.8\linewidth}{!}{
    \begin{tabular}{c|cccc}
    \hline
     Model &FID $\downarrow$ &LPIPS$\downarrow$ &DISTS$\downarrow$ &PSNR$\uparrow$  \\ 
    \hline
    Variant 1 &{\red{$\bm{49.83}$}} &\red{$\bm{0.103}$} &\red{$\bm{0.084}$} &\red{$\bm{26.88}$} \\
    Variant 3 &$61.36$ &$0.113$ &$0.095$ &$26.14$ \\
    Variant 4 &\blue{$58.36$} &\blue{$0.111$} &\blue{$0.092$} &\blue{$26.30$} \\
    \hline
    \end{tabular}
    }
    \caption{Compared to our proposed GPP-LN, removing the global perceptual score or directly adding it to $\mbf{\hat{z}}^{T}_{nl}$ lead to quite poor results.}
    \label{tab_ab_global}
    \end{subtable}
\caption{Ablation results on LOL dataset. Both perceptual priors and our proposed GPP-LN and LPP-Attn have positive contributions to LLIE. [Key:\textbf{\red{Best}}, \blue{Second Best}]}    
\label{table_abla}
\end{table}

%% file: 0_Sections/6_conclusions.tex
\section{Conclusion}
\label{conclusion}

To achieve adaptive and realistic enhancement in real-world scenarios, we introduce a novel LLIE framework (\textbf{GPP-LLIE}) guided by generative perceptual priors. We firstly develop a pipeline for generative perceptual priors extraction based on pre-trained VLMs. Specifically, we design several text prompts to guide VLMs to assess low-level attributes of low-light images globally and locally. We also introduce a sigmoid-based quantification strategy to output the global score and local quality map. Moreover, we introduce a transformer network as the backbone for the diffusion process, where we design the global perceptual priors modulated layer normalization and local perceptual priors guided attention mechanism to guide the enhancement. We evaluate our method on $3$ paired and $4$ real-world datasets, demonstrating superior performance and good generalization.